\documentclass[letterpaper, 10 pt, conference]{ieeeconf}  

\IEEEoverridecommandlockouts                              

\usepackage{cite}
\usepackage{amsmath,amssymb,amsfonts}
\usepackage{graphicx}
\usepackage{textcomp}
\usepackage{xcolor}
\usepackage{soul, color, xcolor}
\usepackage{ulem}
\usepackage{soul}
\usepackage{hyperref}
\usepackage{svg}
\usepackage{multirow}
\usepackage{subcaption}
\usepackage{algorithm}
\usepackage{algpseudocode}
\usepackage{CJKutf8}
\usepackage{array}
\usepackage{tabularray}
\def\BibTeX{{\rm B\kern-.05em{\sc i\kern-.025em b}\kern-.08em
    T\kern-.1667em\lower.7ex\hbox{E}\kern-.125emX}}
\graphicspath{ {image/} }

\title{\LARGE \bf
\begin{CJK*}{UTF8}{gbsn}  
Decoding RobKiNet: Insights into Efficient Training of Robotic Kinematics Informed Neural Network
\end{CJK*}
}

\author{Yanlong Peng$^{1}$, Zhigang Wang$^{2}$, Ziwen He$^{1}$, Pengxu Chang$^{1}$, Chuangchuang Zhou$^{3}$,  Yu Yan$^{4}$, Ming Chen$^{1}$*
\thanks{This work was supported by the Ministry of Industry and Information Technology of China for financing this research within the program "2021 High Quality Development Project (TC210H02C)"}
\thanks{$^{1}$ School of Mechanical Engineering, Shanghai Jiao Tong University, Shanghai, China. (e-mail: \{me-pengyanlong, heziwen, 19119175490, mingchen\}@sjtu.edu.cn)}%
\thanks{$^{2}$ Intel Labs China, Beijing, China. (e-mail: zhi.gang.wang@intel.com)}%
\thanks{$^{3}$ Henan Academy of Sciences, Zhengzhou, China. (e-mail: chuangchuang.zhou@hnas.ac.cn)}%
\thanks{$^{4}$ Intel CCG FIS, Shanghai, China. (e-mail: yu.yan@intel.com)}%
}

\begin{document}
\begin{CJK*}{UTF8}{gbsn}

\maketitle
\thispagestyle{empty}
\pagestyle{empty}

\begin{abstract}
In robots task and motion planning (TAMP), it is crucial to sample within the robot's configuration space to meet task-level global constraints and enhance the efficiency of subsequent motion planning.
Due to the complexity of joint configuration sampling under multi-level constraints, traditional methods often lack efficiency. This paper introduces the principle of RobKiNet, a kinematics-informed neural network, for end-to-end sampling within the Continuous Feasible Set (CFS) under multiple constraints in configuration space, establishing its Optimization Expectation Model. Comparisons with traditional sampling and learning-based approaches reveal that RobKiNet’s kinematic knowledge infusion enhances training efficiency by ensuring stable and accurate gradient optimization.
Visualizations and quantitative analyses in a 2-DOF space validate its theoretical efficiency, while its application on a 9-DOF autonomous mobile manipulator robot(AMMR) demonstrates superior whole-body and decoupled control, excelling in battery disassembly tasks. RobKiNet outperforms deep reinforcement learning with a training speed 74.29 times faster and a sampling accuracy of up to 99.25\%, achieving a 97.33\% task completion rate in real-world scenarios.

\end{abstract}

\section{INTRODUCTION}
\begin{figure}[t]
    \centering
    \includegraphics[scale=0.21]{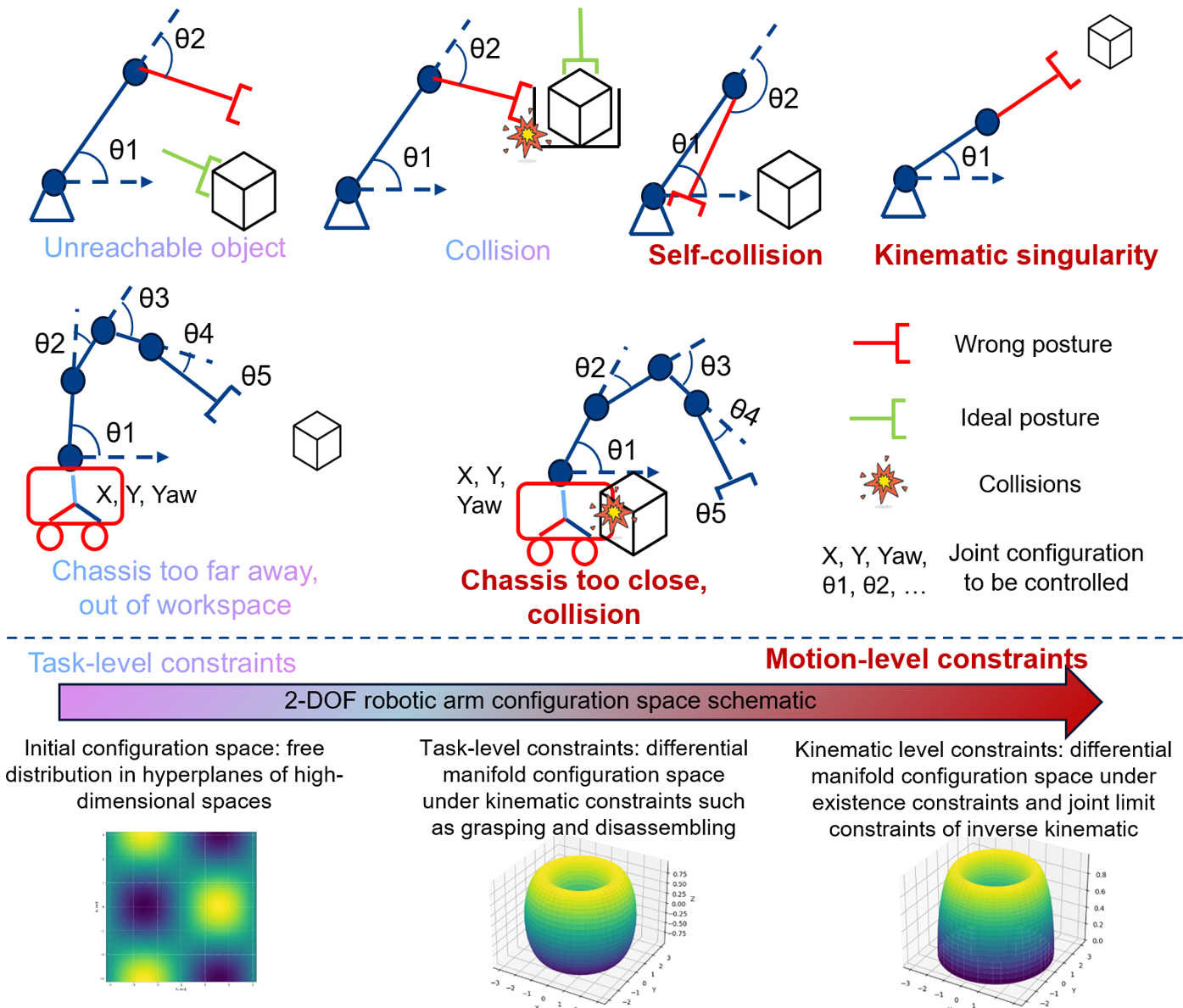}
    \caption{Formation of the Constrained CFS. In robotic TAMP, there are multiple constraints at both the task and motion levels. It is necessary to select appropriate joint configurations from the configuration space to satisfy these combined constraints. (Top part): Errors in sampling joint configurations for robots with different degrees of freedom can lead to violations of constraints at various levels. (Bottom part): The imposition of constraints from the task level to the motion level progressively reduces the configuration solution space, with the ideal configuration space being the CFS.}
    \vspace{-5mm}
    \label{FIG: CFS define.png}
\end{figure}

Robot task and motion planning (TAMP) \cite{guo2023recent, munawar2018maestrob, garrett2021integrated} needs to focus on high-level goals of robot-environment interaction, such as target position and grasping direction, within the constraints at the task level. Meanwhile, the motion-level constraints focus on the lower-level limitations during the actual execution of the task, such as kinematic constraints and joint limits \cite{guo2023recent, bouhsain2023learning}. The inclusion of multi-level and complex constraints greatly reduces the feasible solution space of robot configurations, which is often a high-dimensional differential manifold structure \cite{kim2021learning}. Therefore, how to effectively sample configuration parameters within the Continuous Feasible Set(CFS) under multi-level constraints \cite{qureshi2020neural} to satisfy the task and motion-level constraints is a critical challenge (Figure \ref{FIG: CFS define.png}) .

The feasible solution space $\mathcal{C}_{\text{CFS}}$ forms a subspace of the original configuration $\mathcal{C}$. This subspace, referred to as the CFS, can be defined as:
\begin{equation}\label{Equ:CFSdefine}
\mathcal{C}_{\text{CFS}} = \left\{ {\theta} \in \mathcal{C} \mid g_i({\theta}) = 0, \, h_j({\theta}) \leq 0, \forall i, j \right\}
\end{equation}
where $\theta$ represents the joint configuration. $g_i$ denotes equality constraints, such as task-level position and orientation constraints. $h_j$ denotes inequality constraints, such as motion-level joint limits constraints.

Currently, mature sampling-based algorithms seek feasible solutions through random sampling(RS) within the CFS \cite{kingston2018sampling}. Some TAMP problems can be converted into convex optimization problems, with specific solvers implemented to ensure constraint satisfaction \cite{liu2018convex}. However, these methods often struggle to guarantee the global optimal solution and exhibit low efficiency when dealing with high-dimensional task constraints and stringent motion limitations. Learning-based approaches are also widely used. Deep reinforcement learning, in essence, learns a policy to find the configuration parameters with the highest reward within the CFS \cite{9699963}, but training the policy is complex. Vision-Language-Action(VLA) agents directly bridge the gap in multi-level constraint discrete planning \cite{ma2024survey}, outputting feasible actions. However, this faces challenges such as sparse real operational data, making it difficult to efficiently train the model.

In the face of multiple levels of constraints in TAMP problems, the challenge of effectively sampling configuration parameters within the CFS has been addressed in our previous work through the Robotic Kinematics Informed Neural Network (RobKiNet) \cite{RobKiNet}. RobKiNet injects prior knowledge—such as the forward kinematics (FK) and inverse kinematics (IK) constraints of robotic arms in arbitrary dimensions, joint angle limits, etc.—into the neural network training process, enabling direct end-to-end sampling of configuration parameter solutions within the CFS. This paper further delves into the RobKiNet approach, focusing on the fundamental reasons behind its efficient training from a theoretical perspective. The injection of constraints like kinematics during the training phase directly facilitates the convergence of the end-to-end network to the solution space from task-level constraints to motion-level constraints, which corresponds to manifold embedding in configuration space \cite{gu2000configuration}. This knowledge-guided approach brings significant benefits: the differential manifold solution space contracts rapidly, resulting in high data efficiency. Therefore, when TAMP planning is required, task-level commands (such as the ideal pose constraints for the end-effector) can be directly mapped to motion-level joint configurations through RobKiNet.


The main contributions of this paper are as follows:
\begin{enumerate}
\item Developed a probabilistic Optimization Expectation Model to explain RobKiNet’s convergence in CFS sampling under multi-level constraints, validated against traditional methods.
\item Demonstrated that kinematic prior knowledge ensures stable and accurate gradient optimization, with quantitative visualization in a 2-DOF space.
\item Validated RobKiNet on a 9-DOF AMMR, achieving a 74.29× training speed improvement over deep deterministic policy gradient(DDPG) and a 97.33\% task completion rate in real-world battery disassembly.
\end{enumerate}




\section{Related Work}

\subsection{CFS Problem under Compound Constraints in TAMP}
In many real-world scenarios, the characteristics of TAMP problems involve high-dimensional, non-convex, and composite constraint combinations \cite{wells2019learning, liu2018convex}. These constraints are often due to the need to consider the robot's physical limitations and environmental factors, such as collisions, kinematic constraints, and dynamic behaviors \cite{hanheide2017robot, bouhsain2023learning}.

The sampling of robot configurations is predicated on identifying the CFS, as defined in Equation \ref{Equ:CFSdefine}, which represents the continuous feasible configuration space of the robot under the given constraints \cite{kim2021learning, qureshi2020neural, li2021method, gu2000configuration}. The CFS is constrained not only by the robot's mechanical limitations but also by task requirements, such as maintaining contact forces, avoiding obstacles, and meeting task-specific conditions \cite{stilman2010global}.
As the constraints increase, the feasible region of the CFS shrinks, manifesting as a non-convex high-dimensional manifold embedding, which presents a particularly challenging scenario \cite{gu2000configuration}.

\subsection{Robot Motion Strategies under Multiple Constraints}\label{RELATED WORK B}



Both sampling-based and learning-based strategies attempt to find the ideal solution from the CFS.
Random sampling \cite{pddlstream} and effective bias sampling \cite{kingston2018sampling} are widely used methods in sampling strategies. The core of these methods is to use solvers or simulators to validate the effectiveness of sampled points, with the hope that a limited search will yield feasible solutions within the CFS \cite{pan2015efficient}. 

Among learning-based strategies, deep reinforcement learning methods such as DDPG \cite{DDPG, 9699963, haarnoja2024learning} train agents to learn motion policies, where the agent provides the optimal response under constraints. However, policy development relies heavily on extensive techniques and training resources, often resulting in low efficiency.
Traditional artificial neural networks (ANNs) face challenges such as the long-tail effect of data and non-convex optimization in supervised training \cite{zhou2022dynamically, zhou2018deep}, making it difficult to establish efficient robot motion neural networks.
Recently, Vision-Language-Action (VLA) models \cite{ding2024quar, yue2025deer, ma2024survey} have further integrated multimodal information processing, such as visual inputs and language descriptions, to handle more complex task constraints and environmental conditions, and select the most appropriate configuration parameters within the CFS \cite{yue2025deer, gbagbe2024bi}. However, the difficulty and cost of robot action sample collection, as well as hallucination issues, are key limitations \cite{ma2024survey}.

Physics-Informed Neural Networks introduce more deterministic knowledge representations into the training mode and framework, capturing relationships between network nodes \cite{PINNCASE}. These relationships are often formulated as partial differential equations(PDEs) or boundary constraints derived from human experience or natural language \cite{9403414}. 
This provides some insights for situations like robot kinematics, where there is no closed-form analytical solution.





\section{METHODOLOGY}



In this section, we define the sampling problem from CFS as an optimization problem (Section \ref{3-0}). We then describe the architecture of RobKiNet, establish the Optimization Expectation Model for various methods(Section \ref{3-1}). Following this, we use a 2-DOF robotic arm to visualize and deeply analyze the entire training process of RobKiNet, which confirms its efficiency, as discussed in Section \ref{3-2}. Using this approach, we introduce the method for establishing RobKiNet on a higher-degree-of-freedom robot in Section \ref{3-3}, and the results will be presented in Section \ref{4} .

\subsection{Expectation Modeling}\label{3-0}
In robot TAMP, the convergence of feasible solution spaces within the configuration space fundamentally results from the accumulation and transfer of task-level constraints to motion-level constraints, which leads to the dimensionality reduction of the continuous high-dimensional solution space (e.g., manifold embedding).

Given a desired target pose $pose_{\text{target}}$, the problem of determining a valid joint configuration within the CFS can be formulated as finding a mapping function $\mathcal{F}$, such that:
\begin{equation}
\mathcal{F}(pose_{\text{target}}) = {\theta}, \quad \text{where } {\theta} \in \mathcal{C}_{\text{CFS}}.
\end{equation}

Due to the topological properties of differential manifolds, the joint configuration space is locally homeomorphic to Euclidean space. Therefore, finding mapping function $\mathcal{F}$ can be defined as an optimization problem. Using an n-DOF robotic arm as an example, if a set of joint angles within the constrained CFS is computed through motion-level constraints (kinematics) and approaches the task-level constraint (target pose), we can represent the likelihood of this occurrence using a conditional probability distribution $\mathbb{P}$. Our objective is to optimize this probability distribution:
\begin{equation}
\begin{aligned}
    \mathbb{P}&(\theta_1, \cdots, \theta_n | pose_{\text{target}}) \propto \\ & \exp\left( -\frac{\| f(\theta_1, \cdots, \theta_n) - (pose_{\text{target}}) \|^2}{\sigma^2} \right).
\end{aligned}
\end{equation}



This is a Gaussian-like distribution, where $\theta_n$ is the predicted configuration of the $n$-th joint, and $f$ denotes the kinematic computation, which in traditional methods such as random sampling is represented by a physical simulation engine.
Therefore, our expected error loss can be expressed as:
\begin{equation}
    \mathbb{E}_{(\theta_1, \cdots, \theta_n) \sim \mathbb{P}(\theta_1, \cdots, \theta_n | pose_{\text{target}})} \left[ \| f(\theta_1, \cdots, \theta_n) - (pose_{\text{target}}) \|^2 \right].
\end{equation}

For the continuity of the CFS, the solution of the configuration space comes from the differential manifold $Q$, its integral form is expressed as:
\begin{equation}
\begin{aligned}
    \int_{(\theta_1, \cdots, \theta_n) \in Q} & \| f(\theta_1, \cdots, \theta_n) - (pose_{\text{target}}) \|^2 \, \\ & \mathbb{P}(\theta_1, \cdots, \theta_n | pose_{\text{target}}) \, d\theta_1  \cdots d\theta_2.
\end{aligned}
\end{equation}

The ultimate goal of the probability distribution is to prioritize the sampling of the solution space through weights, ensuring that the optimal solutions are concentrated in the CFS that satisfies task constraints, kinematic constraints, and angular constraints.

\subsection{Architecture and Optimization Expectation Model of RobKiNet}\label{3-1}



\begin{figure}[t]
    \centering
    \includegraphics[scale=0.075]{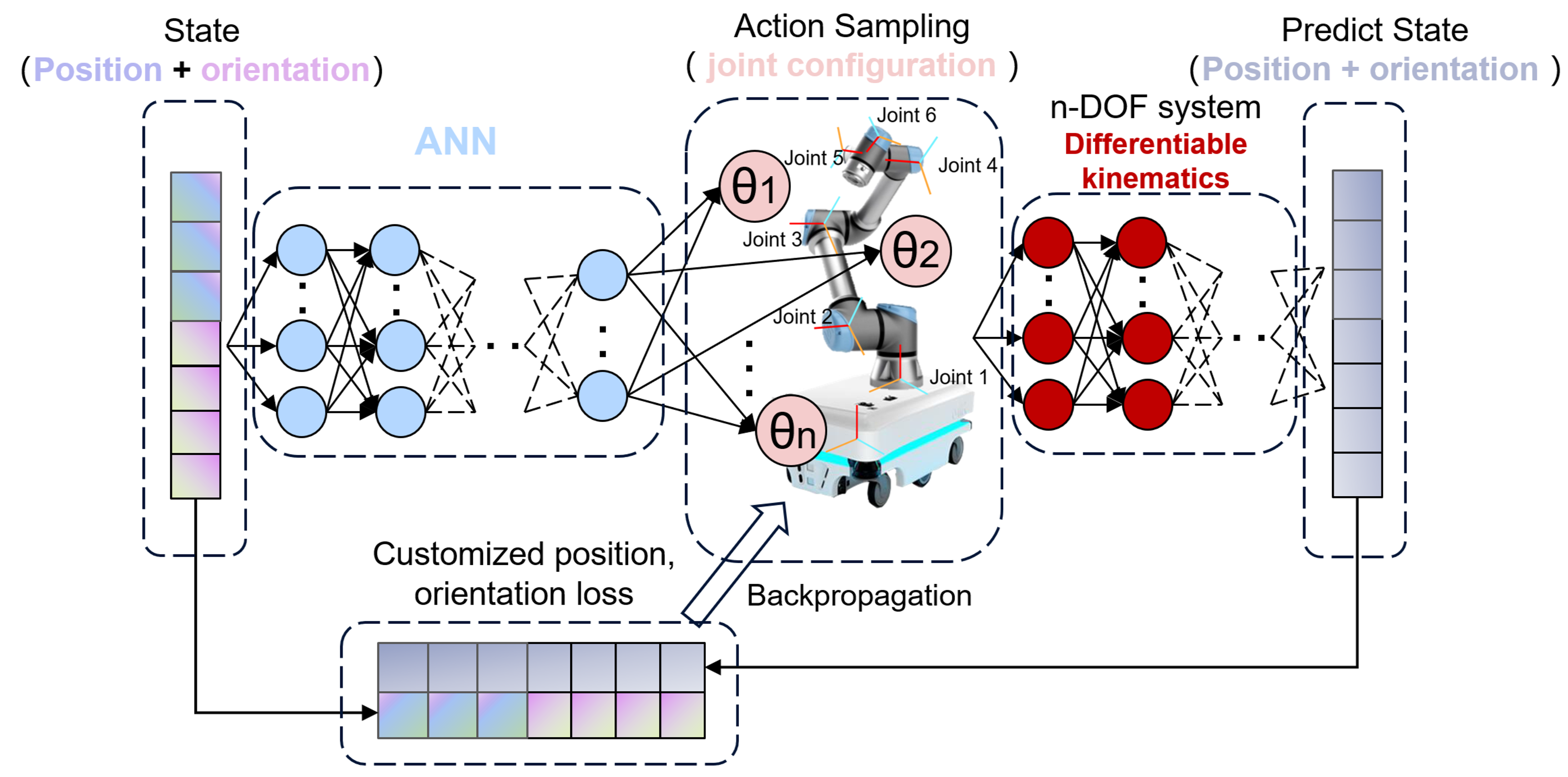}
    \caption{Description of the RobKiNet architecture.
Kinematic constraints are injected into the training process as a priori knowledge (rednodes) and participate in the forward propagation. Task-level constraints (input pose) and motion-level constraints (kinematics) are guaranteed in the backpropagation process through customization loss.}
    \vspace{-5mm}
    \label{FIG: RobKiNet Architecture.png}
\end{figure}

As shown in Figure \ref{FIG: RobKiNet Architecture.png}, the key idea behind RobKiNet is to use differential programming methods \cite{Differentiableprogramming2} to incorporate the FK or IK of the n-DOF system into the forward propagation process. 
It will substantially expand the computational graph of forward propagation in the traditional model, integrating motion-level constraints into the training process as prior knowledge.
For task-level constraints inputs, the joint configurations are directly predicted through an ANN. The predicted results are then fed into the differentiable kinematic computation engine (shown in red in Figure \ref{FIG: RobKiNet Architecture.png}, motion layer constraints) to obtain the predicted pose state. This is compared with the input to calculate the loss, which guides the optimization direction during the backpropagation of the network.

\begin{figure*}[t]
    \centering
    \includegraphics[scale=0.29]{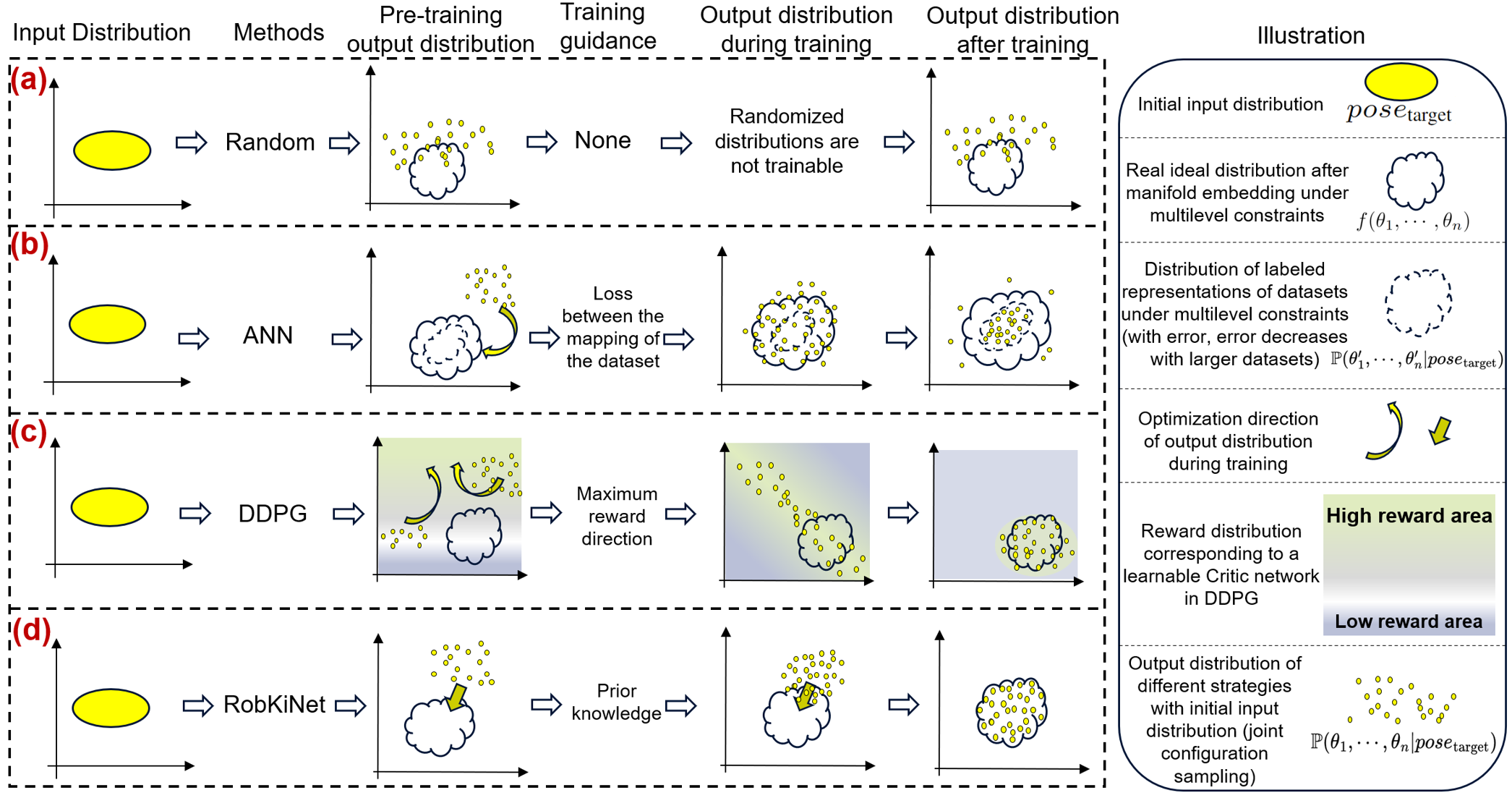}
    \caption{
    Comparison of training processes for different algorithms. The training process involves optimizing their respective probability distribution processes. (a) RS method: The distribution is entirely random due to its untrainable nature. (b) ANN method: The distribution is gradually optimized towards the range of the dataset distribution, but inherently limited by the dataset itself. (c) Deep reinforcement learning method, using DDPG as an example: The output distribution moves towards regions of higher reward. (d) Efficient training of RobKiNet: Knowledge directly guides the distribution.
    }
    \vspace{-4mm}
    \label{FIG: Comparison.png}
\end{figure*}

The methods mentioned in Section \ref{RELATED WORK B} aim to find and optimize the probability distribution to minimize the expected error loss. We built the Optimization Expectation Model for multiple methods based on the content in Section \ref{3-0}. The optimization process is illustrated in Figure \ref{FIG: Comparison.png}.

\noindent \textbf{Random Sampling (Figure \ref{FIG: Comparison.png}(a))}
The sampling results typically require validation through a simulation engine to check whether kinematic and other constraints are satisfied. Resampling occurs if not satisfied. The sampling region corresponds to the configuration space $Q$, and the expected error loss for this distribution is as follows. $Q$ serves as a strategy and is non-optimizable through learning.
\begin{equation}
    \mathbb{E}_{(\theta_1, \cdots, \theta_n) \in Q} \left[ \| f(\theta_1, \cdots, \theta_n) - (pose_{\text{target}}) \|^2 \right].
\end{equation}

\noindent \textbf{Supervised Learning ANN (Figure \ref{FIG: Comparison.png}(b))}
Assume that the known dataset is represented as $((pose_{\text{target}}), (\theta_{1}^{\prime}, \cdots, \theta_{n}^{\prime}))$. Due to the long-tail effect \cite{zhou2018deep}, the dataset distribution is smaller than the true distribution range. The optimization process can be expressed as below where $(\theta_1, \cdots, \theta_n) \sim \mathbb{P}(\theta_1, \cdots, \theta_n | pose_{\text{target}})$.
\begin{equation}
    \mathbb{E} \left[ \| (\theta_{1}^{\prime}, \cdots, \theta_{n}^{\prime}) - (\theta_1,  \cdots, \theta_n ) \|^2 \right].
\end{equation}
The neural network parameters form a mapping ${\mathbb{P}}$. The expected loss from sampling this distribution will be minimized when it is closest to the dataset distribution ${\mathbb{P}(\theta_{1}^{\prime}, \cdots, \theta_n^{\prime} | pose_{\text{target}})}$. This process can be realized through training on a large dataset, but after training, if inputs outside the dataset are encountered, the output will deviate from the true distribution range.

\begin{figure}[t]
    \centering
    \includegraphics[scale=0.28]{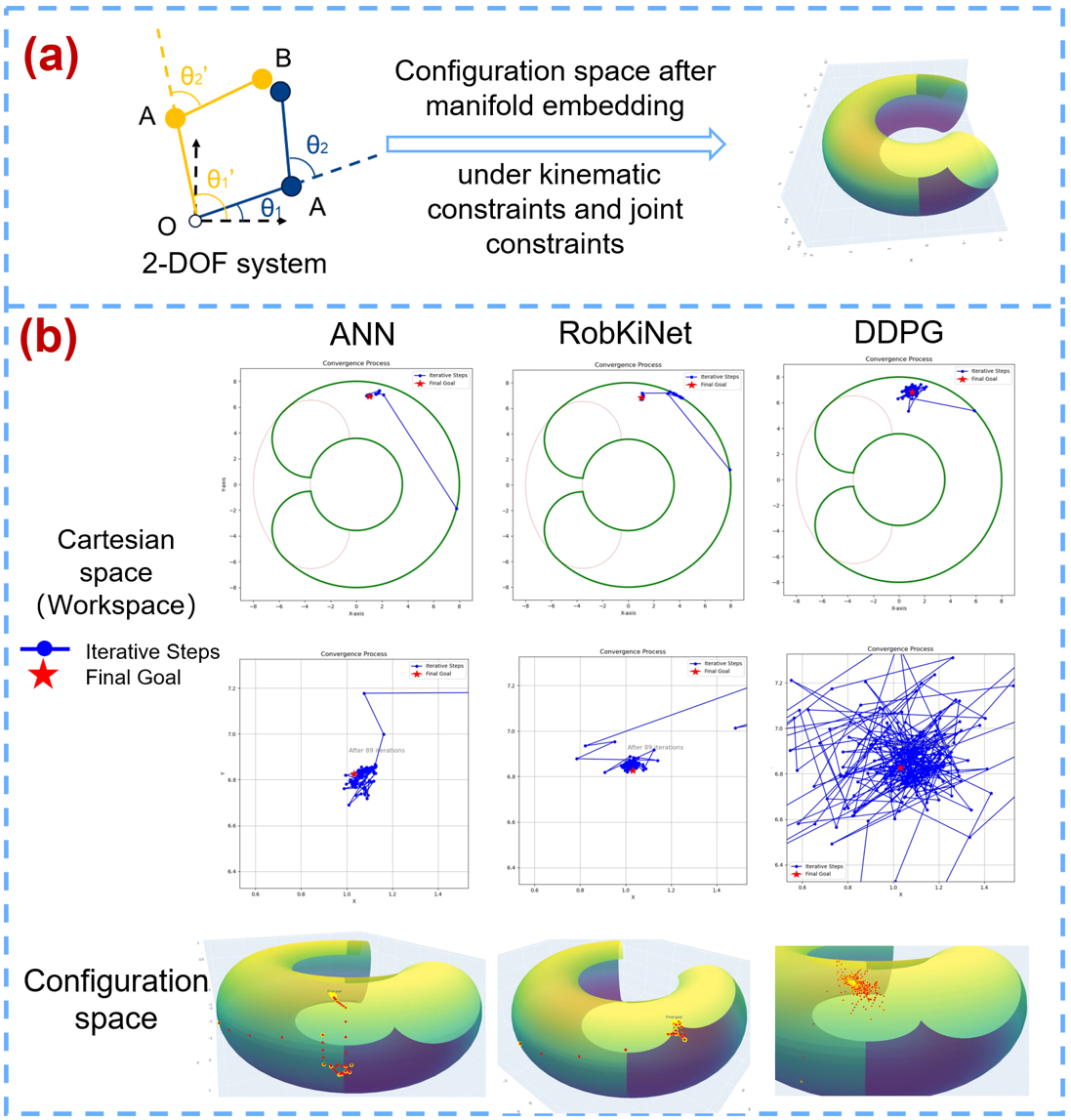}
    \caption{Visualization of the training convergence process of configuration parameter sampling in the CFS for a 2-DOF planar robotic arm, under both task and motion constraints, using different methods. (a) The 2-DOF kinematics have two solutions, corresponding to the left-handed and right-handed coordinate systems. Under the constraints, an ideal solution space manifold embedding occurs. (b) The configurations output by the three methods during the training process after kinematic calculations (motion constraints). The configurations gradually iterate toward the target pose (task constraints).}
    \vspace{-5mm}
    \label{FIG: 2DOF.png}
\end{figure}

\noindent \textbf{DDPG (Figure \ref{FIG: Comparison.png}(c))}
In the deep reinforcement learning optimization process under the Actor-Critic (AC) architecture, the training of the Actor network is guided by the search for the largest reward region, while the reward value $Q$ of the Critic network is also continuously optimized. The training objective comes from the contents of the Buffer under a deterministic policy. The training guidance for the Actor network can be seen as:
\begin{equation}
\label{Actor network}
\begin{aligned}
    &\mathbb{E}_{(\theta_1, \cdots, \theta_n) \sim \mathbb{P}(\theta | \text{policy})} \left[ \nabla_{\theta_1, \cdots, \theta_n} Q(\text{pose}_{\text{target}}, \theta_1, \cdots, \theta_n) \right]
\end{aligned}
\end{equation}
Since the reward of the Critic network is optimized, high-reward regions are constantly shifting, which hinders the training efficiency of the network.

\begin{figure*}[t]
    \centering
    \includegraphics[scale=0.36]{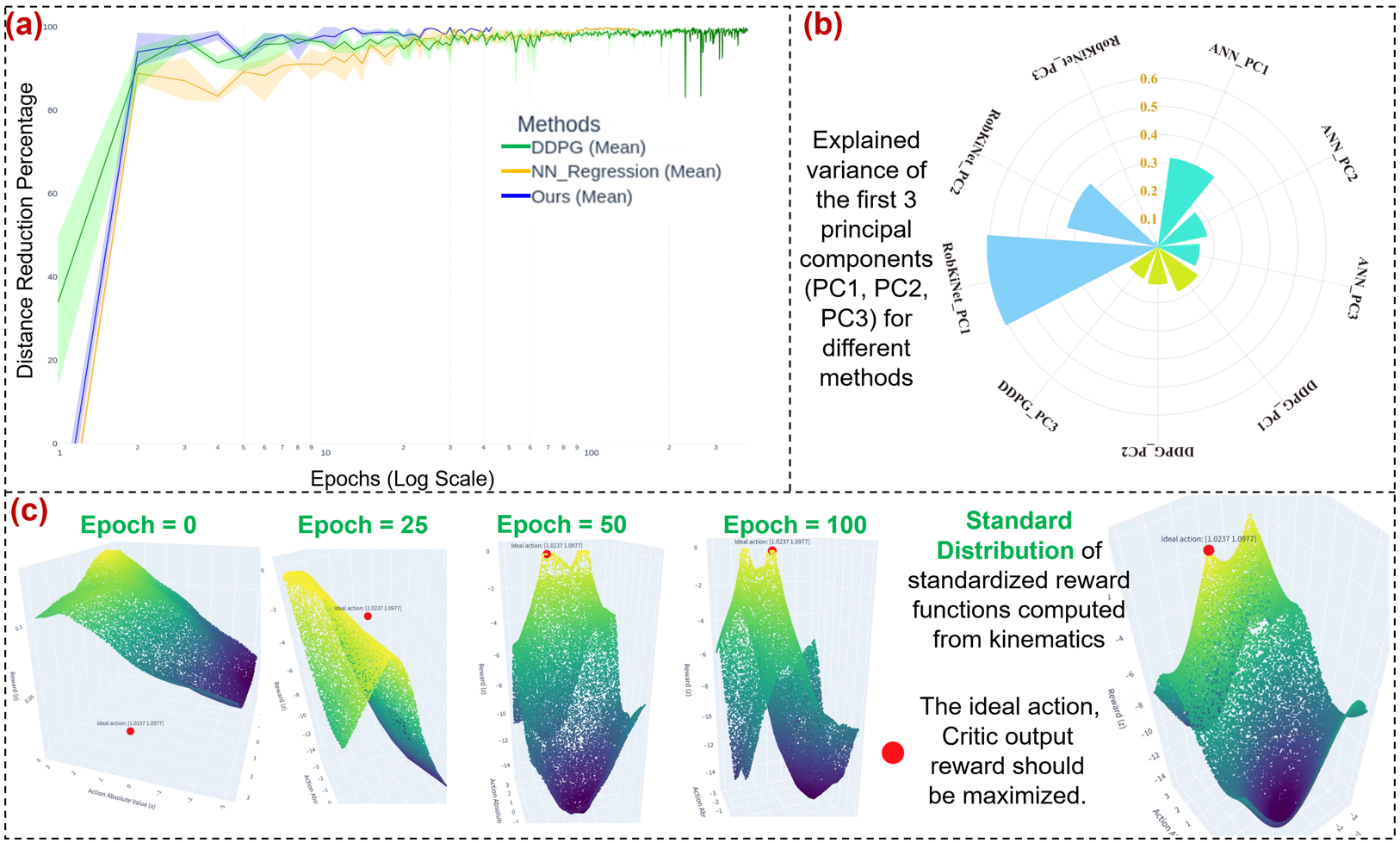}
    \caption{Quantitative analysis of the efficient training process of RobKiNet. (a) Distance Reduction Percentage(DRP) during the iterative training process of different methods. (b) Principal component variance explanation of the gradient vectors for all parameters of the neural network during training. (c) The variation in the distribution of the reward function value guiding the network training in the AC architecture throughout the training process.}
    \vspace{-5mm}
    \label{FIG: quantitative analysis.png}
\end{figure*}

\noindent \textbf{RobKiNet (Figure \ref{FIG: Comparison.png}(d))}
Before training begins, the distribution of the neural network formed by the ANN, $\mathbb{P}(\theta_1, \cdots, \theta_n | \text{pose}_{\text{target}})$, is also a meaningless distribution. The training guidance is provided by the FK and IK calculations incorporated into the network via differential programming, which manifest as a prior distribution under motion-level constraints, $f(\theta_1, \cdots, \theta_n)$. By differentiating $f$, the distance between the predicted points and the task constraint pose can be directly reduced to the ideal distribution.
\begin{equation}\label{the efficient optimization process of RobKiNet}
    \mathbb{E}_{(\theta_1, \cdots, \theta_n) \sim \mathbb{P}(\theta_1, \cdots, \theta_n | \text{pose}_{\text{target}})} \left[ \| f(\theta_1, \cdots, \theta_n) - \text{pose}_{\text{target}}) \|^2 \right].
\end{equation}
Therefore, the efficient optimization process of RobKiNet can be explained as:
\begin{enumerate}
\item The gradient descent is rapid and not influenced by dataset labels, with gradient optimization exhibiting a ‘Stable direction’.
\item Knowledge constrains the solution space with clear "rewards and penalties", and the ideal distribution is well-defined, thus enabling ‘Accurate direction’ in the optimization.
\end{enumerate}

\subsection{Visualization and Quantification of RobKiNet Training for 2-DOF Systems}\label{3-2}

To more intuitively demonstrate the efficient training process of RobKiNet, we consider the training process of a 2-DOF planar robotic arm, as shown in Figure \ref{FIG: 2DOF.png}. The forward kinematics mapping (represented as $f(\theta_1, \cdots, \theta_n)$ in the previous formula) can be expressed as follows, where $l_1$ and $l_2$ denote the lengths of the corresponding manipulator links.
 \begin{equation}
 \begin{aligned}
    \quad f(\theta_1, \theta_2) = \begin{bmatrix} x \\ y \end{bmatrix} = \begin{bmatrix} l_1 \cos \theta_1 + l_2 \cos (\theta_1 + \theta_2) \\ l_1 \sin \theta_1 + l_2 \sin (\theta_1 + \theta_2) \end{bmatrix}. \\
    \theta_1 \in [\theta_{1,\text{min}}, \theta_{1,\text{max}}] \quad \text{and} \quad \theta_2 \in [\theta_{2,\text{min}}, \theta_{2,\text{max}}].
\end{aligned}
\end{equation}

It is clear that the inverse kinematics constraints (two sets of solutions, left-handed and right-handed coordinate systems) and joint limit constraints are the primary motion-level constraints (Figure \ref{FIG: 2DOF.png}(a)). Given an input of an ideal pose (task-level constraint), the training output of different methods iteratively converges toward the ideal target, as shown in Figure \ref{FIG: 2DOF.png}(b). Qualitative analysis reveals that the ANN, guided by the unclear left-handed and right-handed solutions in the dataset, experiences instability in the direction during convergence, being pulled in various directions. The DDPG, due to the reward learning process, exhibits an inaccurate convergence direction and oscillates within a certain range. The efficient training process of RobKiNet is characterized by gradient optimization that exhibits a ‘Stable direction’ and ‘Accurate direction’, which is clearly evident. This can be further visualized through quantitative analysis (Figure \ref{FIG: quantitative analysis.png}).

\noindent \textbf{Gradient Optimization: Stable Direction}
For the iterative optimization process of the three methods, the Distance Reduction Percentage (DRP) is computed across all epochs:
\begin{equation}\label{Eq:DRP}
\begin{aligned}
\text{DRP}(\%)& =\left( 1 - \frac{\| f(\theta_1^{(epoch)}, \cdots, \theta_n^{(epoch)}) - \text{pose}_{\text{target}} \|}{\| f(\theta_1^{(0)}, \cdots, \theta_n^{(0)}) - \text{pose}_{\text{target}} \|} \right) 
\end{aligned}
\end{equation}
In this case, $d_{\text{target}}^{(epoch)}$ represents the distance between the network output and the target point at the $epoch$ , while $d_{\text{target}}^{(0)}$ represents the distance at the beginning of training. This metric quantifies the speed of gradient descent for different methods, and the results clearly demonstrate that RobKiNet converges in fewer than 50 epochs, which is significantly lower than the several hundred epochs required by DDPG (Figure \ref{FIG: quantitative analysis.png}(a)).

Furthermore, calculate the Explained Variance of the principal component(PC) of the gradient magnitudes of all network parameters during the backpropagation at each epoch. The gradient information of each layer (including the gradients of weights and biases) is concatenated into a vector:
$\mathbf{g}_k^{(epoch)}$：
\begin{equation}
\mathbf{g}_k^{(epoch)} = \left( \nabla w_k^{(epoch)}, \nabla b_k^{(epoch)} \right)
\end{equation}
Here, $\nabla w_k^{(epoch)}$ and $\nabla b_k^{(epoch)}$ represent the gradients of the weights and biases of the $k$-th layer at the $epoch$.
By concatenating the gradient information of each layer, the total gradient matrix $\mathbf{G}^{(epoch)}$ is obtained:
\begin{equation}
\mathbf{G}^{(epoch)} = \left[ \mathbf{g}_1^{(epoch)}, \mathbf{g}_2^{(epoch)}, \dots, \mathbf{g}_L^{(epoch)} \right]
\end{equation}
The explanation rate $\lambda_i$ of each principal component from the three methods is calculated and shown(Figure \ref{FIG: quantitative analysis.png}(b)):
\begin{equation}
\begin{aligned}
&\lambda_i = \frac{\text{Var}(\mathbf{PC}_i)}{\sum_{i=1}^{k} \text{Var}(\mathbf{PC}_i)}\quad \\ &\text{where} \quad \mathbf{G}^{(epoch)} \xrightarrow{\text{PCA}} \mathbf{PC}_1, \mathbf{PC}_2, \dots, \mathbf{PC}_k
\end{aligned}
\end{equation}
The high explanation rates of the first few principal components indicate that the direction of the optimization process is stable, and the gradients of a set of core parameters guide the optimization direction.

 \noindent \textbf{Gradient Optimization: Accurate Direction}
For the Actor-Critic architecture in reinforcement learning, Equation \ref{Actor network} illustrates the training guidance for its Actor network. During the training process, the convergence of its Critic network is equally crucial as it determines the distribution of high-reward regions. For DDPG, the training process of its Critic network is expressed as follows:
\begin{equation}
\begin{aligned}
\text{minimize} \ \mathbb{E} \left[ \| Q(\text{pose}_{\text{target}}, \theta_1, \cdots, \theta_n) - r_t \|^2 \right], \\{(\text{pose}_{\text{target}}, \theta_1, \cdots, \theta_n, r_t) \sim \mathbb{D}}
\end{aligned}
\end{equation}
This represents the sampling of state, action, and reward tuples from the Replay Buffer (denoted as $\mathbb{D}$). The objective of optimizing the Critic network is to make the $Q$-value of the current state-action pair close to the reward, $r_t$.

Figure \ref{FIG: quantitative analysis.png}(c) shows the output rewards of the Critic network in DDPG during training for a set of state-action inputs. The distribution of its reward function changes during training, gradually converging to the reward space directly defined by kinematic constraints, which will serve as the direct training guidance for RobKiNet in Equation \ref{the efficient optimization process of RobKiNet}. This indicates that clear prior knowledge constraints can guide the inputs directly and accurately to the most ideal constrained range, without the need to gradually learn the reward distribution and train the policy based on the rewards.










\subsection{RobKiNet in Higher Dimension}\label{3-3}
We conducted the training and deployment of RobKiNet in higher dimensions. The network's input also includes task-level constraints, such as the poses of points that need to be grasped or disassembled. We aim for RobKiNet to output the joint configuration of the composite robot AMMR to enable decoupled control or whole body control. The output joint configurations need to satisfy motion-level constraints such as forward kinematics and inverse kinematics.

Specifically, for a typical AMMR composed of a mobile base and a 6-DOF robotic arm, its overall configuration is:
\begin{equation}\label{FMP9output}
    \theta = 
  [\underbrace{\psi,x,y}_{\theta_{base}}, \underbrace{q_1, q_2, \dots, q_6}_{\theta_{arm}}]^T,\theta  \in \mathbb{R}^9
\end{equation}
For decoupled control(DC), we hope that given the task-level constraint $\text{pose}_\text{target}$, the base configuration $\theta_{base}$ output by RobKiNet can ensure that $\theta_{arm}$ has a Denavit-Hartenbarg(DH) kinematic solution\cite{DH2}: 
\begin{equation}\label{Eq:DC}
\begin{aligned}
\text{for}&\quad \theta_{\text{base}} = \text{RobKiNetDC}(\text{pose}_{\text{target}}) = [\psi, x, y], \\
& \quad \exists \theta_{\text{arm}} \ \text{such that} \ 
\text{pose}_{\text{target}} = \text{FK}(\theta_{\text{base}}, \theta_{\text{arm}}) \quad
\end{aligned}
\end{equation}
For whole body control(WBC), we aim to use RobKiNet to directly provide all joint configuration samples that conform to the legal CFS:
\begin{equation}\label{Eq:WBC}
\begin{aligned}
\text{for}\quad &\theta_{\text{base}},\theta_{\text{arm}} = \text{RobKiNetWBC}(\text{pose}_{\text{target}}) \\ &= [\psi, x, y,q_1, q_2, \dots, q_6], \\
& \text{such that} \ 
\text{pose}_{\text{target}} = \text{FK}(\theta_{\text{base}}, \theta_{\text{arm}}) \quad
\end{aligned}
\end{equation}
Both the 3-DOF and 9-DOF output modes consider the control of the entire AMMR system. The experimental results will be elaborated in detail in the next chapter.

\begin{table}[t]
\caption{Training Effects of Each Method in Three Control Dimension Scenarios}
\vspace{-4mm}
\begin{center}
\renewcommand{\arraystretch}{1.5} 

\begin{tblr}{
  cell{2}{1} = {r=3}{},
  cell{5}{1} = {r=3}{},
  cell{8}{1} = {r=3}{},
  vlines,
  hline{1-2,5,8,11} = {-}{},
  hline{3-4,6-7,9-10} = {2-4}{},
   column{1} = {c}, 
  column{2} = {c},
  column{3} = {c},
  column{4} = {c}
}
\begin{tabular}[c]{@{}c@{}}Control\\dimension  \end{tabular}             & Methods       &\begin{tabular}[c]{@{}c@{}} Number of epochs\\required at 98\% DRP  \end{tabular}  &\begin{tabular}[c]{@{}c@{}} Training epoch\\optimization factor   \end{tabular}        \\
2-DOF            & ANN      & 83.67           & 3.84×           \\
                 & DDPG     & 321.67          & 1×(base)       \\
                 & RobKiNet & 4.33            & \textbf{74.29×} \\
\begin{tabular}[c]{@{}c@{}} 9-DOF\\DC  \end{tabular} & ANN      & unreachable             & /               \\
                 & DDPG     & 6005            & 1×(base)        \\
                 & RobKiNet & 123.2           & \textbf{48.74×} \\
\begin{tabular}[c]{@{}c@{}} 9-DOF\\WBC \end{tabular}& ANN      & unreachable             & /               \\
                 & DDPG     & 29122           & 1×(base)       \\
                 & RobKiNet & 976.33          & \textbf{29.82×} 
\end{tblr}
\vspace{-6mm}
\end{center}\label{Training Effects of Each Method in Three Control Dimension Scenarios}
\end{table}

\section{Experiments and Results}\label{4}
In this section, we will demonstrate the advantages of RobKiNet in the following aspects: (1) training convergence efficiency, (2) deployment test accuracy, and (3) performance in real-world task scenarios. For training efficiency, we will use deep learning and deep reinforcement learning as benchmarks, conducting experiments across different dimensions. Furthermore, we will perform real-world deployment tests during the autonomous movement of an AMMR in the automotive battery disassembly scenario. We compare the sampling accuracy of four methods and set up real task tests for RobKiNet across three different scenarios.

\begin{figure*}[t]
    \centering
    \includegraphics[scale=0.45]{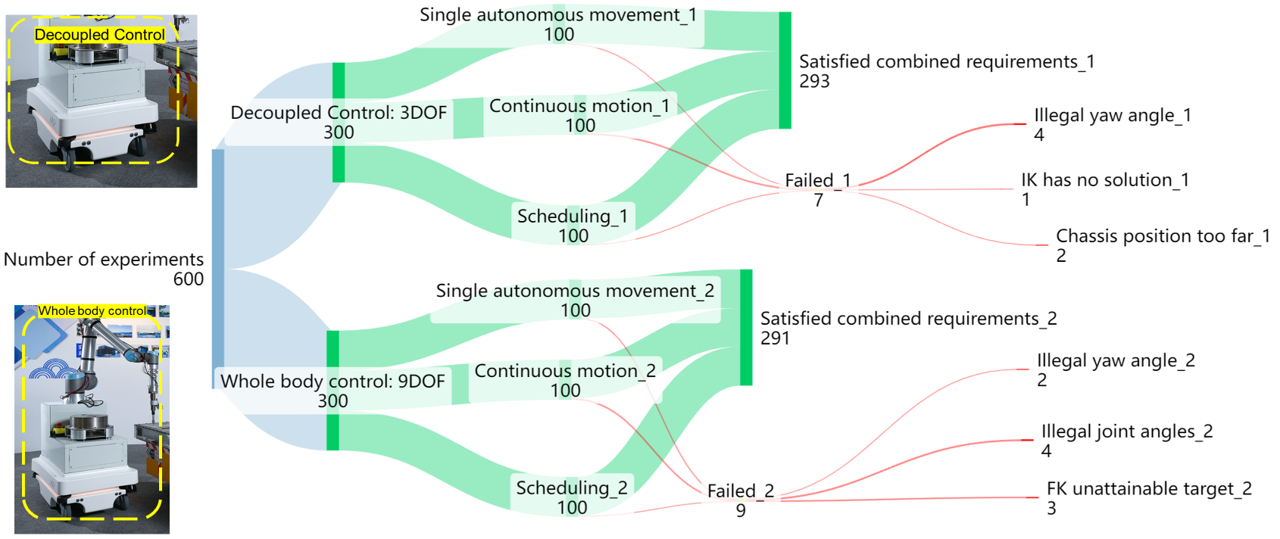}
    \caption{Experimental results of real sampling in a 9-DOF AMMR system using RobKiNet. The experiments were performed in decoupled control and whole-body control modes, with 300 groups in each mode.}
    \vspace{-5mm}
    \label{FIG: Experiment.png}
\end{figure*}

\subsection{Efficient training of RobKiNet}

To validate the efficient convergence of the RobKiNet method, we conducted training using three methods in three scenarios: a 2-DOF planar robot,  whole body control of a 9-DOF AMMR (9-DOF WBC, Equation \ref{Eq:WBC}), and decoupled control of it (9-DOF DC, Equation \ref{Eq:DC}). For the ANN method, we collected 30,000 data sets for tasks in three dimensions. The 2-DOF scenario includes the distribution of left and right hand and angle constraints, while the 9-DOF scenario includes different numbers of DH analytical solutions. For the DDPG method, we used deterministic kinematics in corresponding dimensions as a resampling strategy. The initial data consists of 20,000 sets, with 512 sets resampled each epoch to expand the Buffer. Both RobKiNet and ANN methods employed Ray-tune \cite{tune} to find optimal solutions among three network structures, six different learning rates, and eight batch sizes, training for over 1000 epochs to ensure convergence. We used the DRP (Equation \ref{Eq:DRP}) formula as a quantitative computation standard  to calculate the number of epochs required for each method to converge to 98\% DRP. Using DDPG as the baseline, we calculated the training epoch optimization factor in each scenario. The experimental results are shown in Table \ref{Training Effects of Each Method in Three Control Dimension Scenarios}.


The experiments revealed that, in low-dimensional cases, DDPG's process of finding the optimal strategy is slower, and ANN learns sampling strategies directly from relatively simple dataset mappings, achieving an epoch optimization factor 3.84 times that of DDPG. Clear prior knowledge guidance allows RobKiNet to quickly converge near the accurately constrained poses, achieving an optimization factor 74.29 times that of DDPG. In high-dimensional cases, providing a legal joint configuration that satisfies multi-level constraints in one attempt presents significant convergence difficulty for ANN. While DDPG can achieve this through extensive training, fundamentally due to the comprehensive strategy derived from a large amount of data. In contrast, RobKiNet consistently maintains a significant advantage across all dimensions due to its stable and accurate direction.
\subsection{Real Scenario Deployment Experiment}

To evaluate the performance of RobKiNet in real-world scenarios, we tested the deployment sampling accuracy of four sampling methods across three dimensions. Furthermore, we deployed RobKiNet in an AMMR system \cite{peng2024revolutionizing} for the disassembly of fasteners in automotive battery scenarios.

Table \ref{table:2} presents the sampling accuracy of four sampling methods in different scenarios. Based on the accuracy requirements of real scenarios, we consider a configuration prediction as a positive sample if the Euclidean distance between the predicted and ideal poses is within 1mm. The sampling limit is set at 300 attempts. In all cases, the joint configurations provided by RobKiNet within the CFS were significantly superior to those obtained by random sampling methods and traditional learning-based methods. Compared to DDPG, RobKiNet's accuracy improved by 8.92\% in the 2-DOF scenario and by 25.45\% in the 9-DOF scenario.

\begin{table}[t]
\caption{Real-world Deployment Sampling Accuracy}
\vspace{-4mm}
\begin{center}
\renewcommand{\arraystretch}{1.5} 
\begin{tabular}{|c|c|c|c|} 
\hline
Methods     & 2-DOF   & \begin{tabular}[c]{@{}c@{}} 9-DOF\\DC   \end{tabular} & \begin{tabular}[c]{@{}c@{}}9-DOF\\WBC   \end{tabular}   \\ 
\hline
RS       & 4.65\%  & 3.22\%           & less than 0.33\%            \\ 
\hline
ANN      & 62.20\% & 51.30\%          & less than 5\%               \\ 
\hline
DDPG     & 90.33\% & 77.62\%          & 72.55\%           \\ 
\hline
RobKiNet & 99.25\% & 96.67\%          & 98.40\%           \\
\hline     
\end{tabular}\label{table:2}
\vspace{-7mm}
\end{center}
\end{table}


Additionally, we established three experimental tasks in the battery disassembly scenario. \textbf{Task 1}: Single autonomous movement, which requires the AMMR to complete a movement from a distance and disassemble a single bolt. This task demonstrates the single prediction accuracy of RobKiNet. \textbf{Task 2}: Continuous motion, which requires the AMMR to perform large-range movements and make multiple successive configuration predictions to disassemble more than 15 bolts. This task demonstrates the stability of RobKiNet's output distribution under multiple inputs. \textbf{Task 3}: Scheduling, which requires the AMMR to navigate between multiple areas. This task illustrates RobKiNet's adaptability and generalization when faced with large-range inputs.

As shown in the Figure \ref{FIG: Experiment.png}, when the AMMR performs TAMP, utilizing RobKiNet for joint configuration sampling within the CFS under multiple constraints, the average task completion rate reached 97.33\%.

\section{CONCLUSIONS}


The TAMP problem in robotics often faces the challenge of efficiently sampling configurations within the CFS under multi-level constraints. This work provides an in-depth analysis of the architecture of a kinematic informed neural network, RobKiNet, and establishes an Optimization Expectation Model for different methods addressing the configuration sampling problem within the CFS. The fundamental reason for its efficiency compared to other methods is elucidated: RobKiNet's efficient training under multiple constraints stems from its kinematic infusion, resulting in a stable and accurate direction in gradient optimization. Our visualization and quantitative analysis of the training convergence process in a 2-DOF low-dimensional configuration space strongly support our theoretical assertions. Furthermore, we implemented RobKiNet for whole-body and decoupled control in a 9-DOF AMMR, conducting real-world experiments in the context of battery disassembly. The experiments demonstrated significant advantages of RobKiNet over deep reinforcement learning methods such as DDPG. Its training epoch optimization factor is 74.29 times that of DDPG, with a sampling accuracy reaching up to 99.25\%. In real-world tasks, RobKiNet achieved an average task completion rate of 97.33\%. 

While RobKiNet enables end-to-end control in finite high-dimensional spaces, its applicability to systems lacking precise kinematics (e.g., soft bodies) relies on data-driven learning informed by weak structural or geometric priors. Future efforts will integrate dynamic constraints (e.g., collision avoidance, torque limits) to enhance embodied agents' adaptability in complex environments.


\bibliographystyle{unsrt} 
\bibliography{ref}

\end{CJK*}
\end{document}